\documentclass[review]{elsarticle}

\usepackage{hyperref}
\usepackage{xcolor}

\journal{Journal of \LaTeX\ Templates}

\begin{document}

\begin{frontmatter}

\title{Application of multilayer perceptron with data augmentation in nuclear physics}


\author[mymainaddress]{H\"{u}seyin Bahtiyar}
\ead{huseyin.bahtiyar@msgsu.edu.tr}

\author[mysecondaryaddress]{Derya Soydaner\corref{mycorrespondingauthor}}
\cortext[mycorrespondingauthor]{Corresponding author}
\ead{derya.soydaner@kuleuven.be}

\author[mythirdaddress]{Esra Y\"uksel}
\ead{eyuksel@yildiz.edu.tr}

\address[mymainaddress]{Department of Physics, Mimar Sinan Fine Arts University, Istanbul, Turkey}
\address[mysecondaryaddress]{Department of Brain and Cognition, University of Leuven (KU Leuven), Leuven, Belgium}
\address[mythirdaddress]{Department of Physics, Yildiz Technical University, Istanbul, Turkey}

\begin{abstract}
Neural networks have become popular in many fields of science since they serve as promising, reliable and powerful tools. In this work, we study the effect of data augmentation on the predictive power of neural network models for nuclear physics data. We present two different data augmentation techniques, and we conduct a detailed analysis in terms of different depths, optimizers, activation functions and random seed values to show the success and robustness of the model. Using the experimental uncertainties for data augmentation for the first time, the size of the training data set is artificially boosted and the changes in the root-mean-square error between the model predictions on the test set and the experimental data are investigated. Our results show that the data augmentation decreases the prediction errors, stabilizes the model and prevents overfitting. The extrapolation capabilities of the MLP models are also tested for newly measured nuclei in AME2020 mass table, and it is shown that the predictions are significantly improved by using data augmentation.
\end{abstract}

\begin{keyword}
Deep neural networks \sep nuclear binding energy \sep regression \sep data augmentation
\end{keyword}

\end{frontmatter}


\section{Introduction}

The binding energy or mass is a fundamental property of atomic nuclei, which carries information about the strong interaction between nucleons. While the accurate prediction of nuclear properties (masses, separation energies, radii, etc.) is quite essential to better understand the properties of nuclei and develop nuclear models, these predictions are also used as inputs in the nuclear astrophysical simulations, e.g., r-process nucleosynthesis \cite{APRAHAMIAN2005535,MUMPOWER201686}. To date, many experimental \cite{Wang_2017, Wang_2021} and theoretical studies \cite{PhysRevC.84.014333,WANG2014215,PhysRevLett.108.052501, XIA20181, Erler2012} have been performed to predict nuclear properties as accurately as possible. Presently, we know the properties of around 2500 nuclei from experimental studies \cite{Wang_2017, Wang_2021}, while theoretical model calculations predict that around 7000 nuclei should exist \cite{Erler2012}. The experimental studies are challenging for weakly bound nuclei near the drip lines; hence, the predictions for the nuclear properties mainly rely on the theoretical model calculations. Although the theoretical models are quite successful in predicting nuclear properties along the nuclear stability line, results display a large variety for nuclei near the nuclear drip lines with extreme proton-to-neutron ratio. Therefore, extensive studies are necessary to understand the reasons for these discrepancies and predict the properties of nuclei near the drip lines better as well as finding alternative tools to predict nuclear properties.


Neural networks that date back to the 1940s \cite{McCulloch1943} have grown their impact tremendously in the last decade. Today, we encounter many deep neural network applications in various fields because of the huge amount of data and computational power we have today \cite{krizhevsky2012,brown2020,silver2016}. The gathering of computing power with various kinds of big data has greatly helped the recent rise of deep learning. As a subfield of machine learning, deep neural networks have also been used in the estimation of the various nuclear properties. Since neural networks provide fast and reliable results with much less effort compared to the theoretical nuclear model calculations, they are used as alternative tools to predict the properties of atomic nuclei. To date, a variety of machine learning models have been used either to improve the accuracy of the theoretical model predictions or to predict the nuclear properties by directly using them. One of the pioneering works to predict nuclear properties using neural networks has been conducted in \cite{GAZULA19921}, and then followed by other studies in Refs. \cite{GERNOTH19931, gernoth1995, 10.1007/BFb0104279,ATHANASSOPOULOS2004222, Akkoyun_2013, BAYRAM2014172, PhysRevResearch.2.043363, PhysRevC.80.044332, PhysRevLett.124.162502,Pastore_2021}. In recent years, the application of the neural network models to improve the predictive power of the nuclear models using the available experimental data has also gained considerable interest due to its success in the predictions of several nuclear properties \cite{NIU201848, PhysRevC.98.034318,PhysRevC.93.014311, Utama_2016, PhysRevC.96.044308, PhysRevC.97.014306}.

We know that as the amount of training data increases, neural networks perform better and their generalization ability improves \cite{krizhevsky2012, halevy2009, banko2001}. On the other side, data collection has a cost, or it may not be possible to collect large enough data. Therefore, the insufficient quantity of training data is a real challenge in machine learning \cite{geron2017}. Although we know the properties of many nuclei from the experimental results, these data are not enough to get high accuracy results by training neural network models. An alternative solution to this problem is applying data augmentation techniques. These techniques may be seen as a way of preprocessing the training set only \cite{goodfellow2016}. In machine learning, data augmentation is a well-known technique to increase the training data artificially by applying small modifications to the training data set. In the last decade, this technique was widely applied to improve the model predictions for image data sets \cite{Shorten2019,8388338,8628742}, text data sets \cite{Shorten2021}, and speech data \cite{jaitly2013}. It is shown that data augmentation increases the predictive power, prevents overfitting, and helps to regularize neural network models \cite{goodfellow2016,geron2017}. Since these techniques are mostly used for classification rather than regression problems, their use in the field of regression is a current research area \cite{kuchnik2019,dubost2019,ohno2020}. Considering the scarcity of the available experimental data for some nuclear properties, the application of the data augmentation to the nuclear data sets and investigating the performance of neural networks would be interesting.

In this work, we apply data augmentation techniques to improve the predictive power of neural networks for the prediction of nuclear properties. To increase the number of the training data set, we use the experimental uncertainty in the binding energies of each nucleus. First, we increase the training data by using the experimental error values of each nucleus. Then, we also use the Gaussian augmentation technique to increase training data further. Comparing the results with the available experimental data, the impact of the data augmentation on the predictive power of the neural networks is studied.

The paper is organized as follows. In Sec. II, we summarize the Multilayer Perceptron (MLP) and data augmentation techniques used in this work. In Sec. III, the results are presented. By using different MLP architectures in the calculations, the effect of the data augmentation on the predictive power of the neural networks is discussed. Finally, the conclusions and outlook are given in Sec. IV.

\section{Model: Multilayer Perceptron and Data Augmentation Techniques}

\subsection{Multilayer Perceptron}
Multilayer perceptron (MLP) is a typical kind of feedforward neural network that may include one or more hidden layers between input and output layers. MLP is useful for its ability to solve both classification and regression problems. In this study, our objective is to predict the total nuclear binding energies by using input data. We choose MLP which is mostly preferred in machine learning to solve such regression problems. Considering the success of deep learning in recent years, we also aim to take advantage of deep neural networks by adding multiple hidden layers to MLP.

An example of MLP architecture with two hidden layers is shown in Figure \ref{MLP} where \textit{$\textbf{w}_{1h}$} and \textit{$ \textbf{w}_{2l}$} are the weights belonging to the first and second hidden layers, respectively. In the end, \textit{$\textbf{v}_{l}$} represents the weights belonging to the output layer. These parameters are optimized during training by minimizing a loss function. In our case, MLP takes basic nuclear properties as inputs {\em X}, and predicts the total binding energies as the output {\em Y}. Here, the inputs are the proton (Z) and mass numbers (A) of the relevant nuclei, which define the identity of an atomic nucleus. During training, MLP updates these weights to minimize the difference between the desired output and its predictions.

Basically, training an MLP consists of two parts: {\em Forward} and {\em backward} pass. In the forward pass, the input {\em X} is fed to the input layer. The weighted sum is computed, and the activation propagates in the forward direction for each layer. Each hidden unit usually applies a nonlinear activation function to its weighted sum. Every hidden layer takes the activation of its preceding layer as input. We use ReLU function \cite{glorot2011} which is mostly used for nuclear mass predictions \cite{PhysRevResearch.2.043363}, \cite{yuksel2021} and deep learning in general. In this way, the values of the hidden units in each hidden layer are computed as shown in Eq. \ref{z1} and 2: 

\begin{figure*}[h!]
\centering
\includegraphics[width=\linewidth]{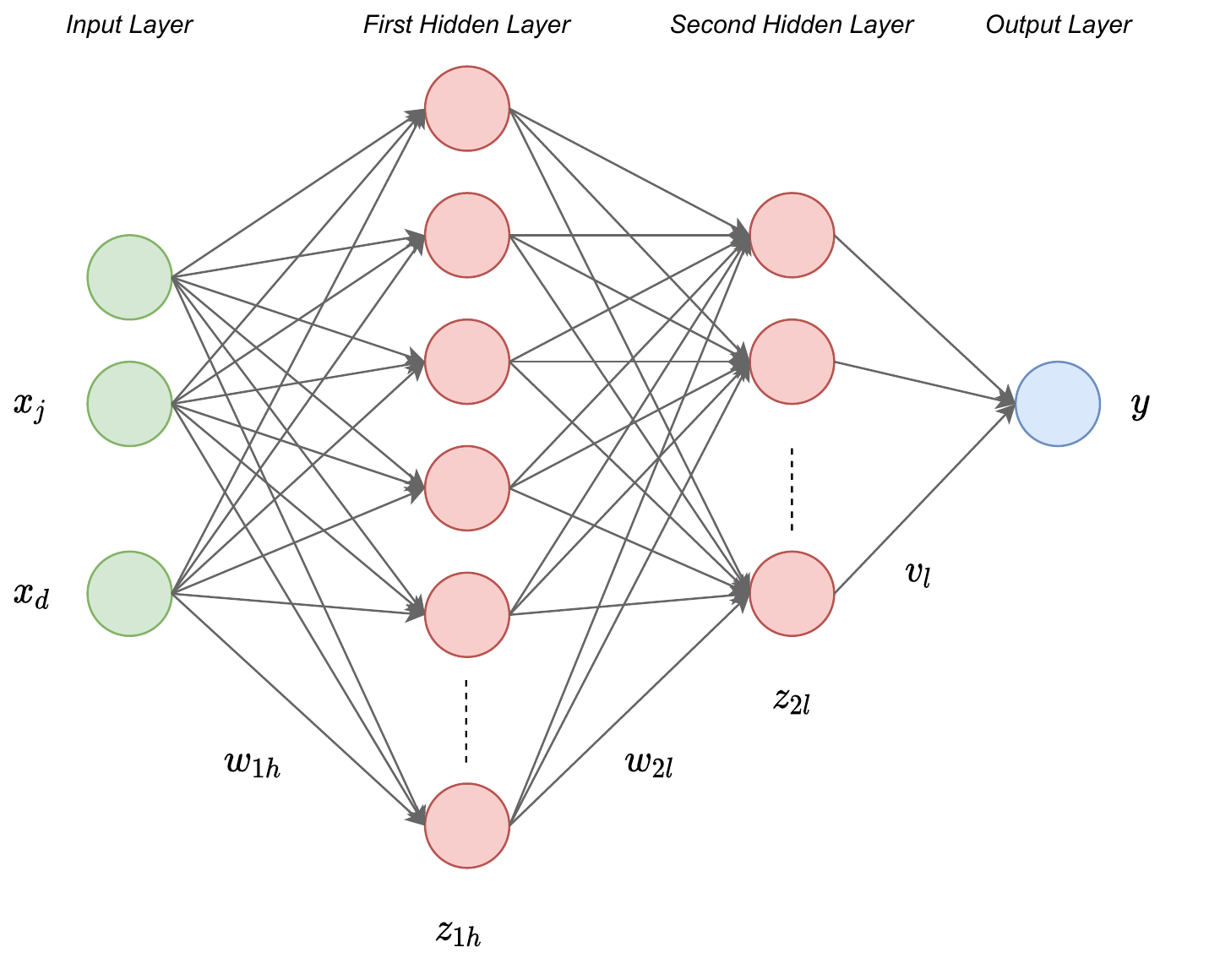}
\caption{The structure of a multilayer perceptron with two hidden layers. In our case, there are two inputs, namely proton (Z) and mass numbers (A) of the relevant nuclei. The binding energy is predicted in the output.}
\label{MLP}
\end{figure*}

\newpage

\begin{eqnarray}
	z_{1h} &=& ReLU( \textbf{w}_{1h}^T \textbf{x}) 
	\nonumber\\
	&=& ReLU\Bigg(\sum_{j=1}^d w_{1hj}x_j + w_{1h0} \Bigg) ,  h=1,...,H_1
	\label{z1}
\end{eqnarray}

\begin{eqnarray}
	z_{2l} &=& ReLU( \textbf{w}_{2l}^T \textbf{z}_1) 
	\nonumber\\
	&=& ReLU\Bigg(\sum_{h=0}^{H_1} w_{2lh}z_{1h} + w_{2l0} \Bigg) ,  l=1,...,H_2
	\label{z2}
\end{eqnarray}

Here, the units of the first and second hidden layers are represented as  \textit{$z_{1h}$} and \textit{$z_{2l}$} \cite{alpaydin2014}. Finally, the output {\em y} is computed by taking the second hidden layer activation \textit{$z_{2}$} as input. As we try to solve a regression problem, there is no non-linearity in the output layer:

\begin{equation}
	y = \textbf{v}^T \textbf{z}_2 = \sum_{l=1}^{H_2} v_{l}z_{2l} + v_{0}
	\label{output}
\end{equation} 

An error is calculated by using a loss function after each output is obtained in the forward pass. In the backward pass, the weights are updated by using this error. We use {\em mean squared error (MSE)} as the loss function of MLP. To train the MLP, we apply the {\em Adam algorithm} \cite{kingma2014} which is mostly preferred in deep learning. The initial learning rate of the Adam algorithm is 0.001, and decay constants are 0.9 and 0.99, respectively. Additionally, we examine the effect of different optimization algorithms in Section \ref{optimizer}. We initialize the layer weights with the {\em Glorot (Xavier) normal initializer} \cite{glorot2010}. More detailed information can be found in \cite{yuksel2021} about MLP and the regression problem as well.

In our experiments, we implement MLP architectures of four different depths. We begin with one single hidden layer, and then add hidden layers one by one until we have four layers. We tried to go deeper than four hidden layers, but we realized that four hidden layers are sufficient for our case. Our MLP architectures are summarized in the first column of Table \ref{table00}. In this column, each number represents the number of hidden units in each hidden layer. In the output layer, only one unit exists as we predict one single numerical value. By examining MLP architectures with different depths, we observe how the total binding energy predictions change with depth.

\subsection{Data Augmentation}
Training a machine learning model on more data is the best way to make it generalize better \cite{goodfellow2016}. Instead of collecting additional data, new training data can be generated from the existing ones using data augmentation techniques. This avoids the cost of collecting new data. Besides, data augmentation is also defined as a regularization technique in machine learning as it reduces overfitting. 

Various data augmentation techniques exist for classification tasks because modifying the training examples with transformations and adding them to the original training data set does not affect the class in the end. For example, in an object recognition task, input images can be slightly shifted or rotated, and the brightness or contrast may be adjusted. Then, the resulting images can be added to the training data set. These transformations do not change the class to which the input images belong. But they force the model to be more tolerant to the position, orientation, and lighting conditions \cite{geron2017}. 

A classifier summarizes a complicated, high-dimensional input with a single category. This means that the main task facing a classifier is to be invariant to a wide variety of transformations \cite{goodfellow2016}. On the other side, in a regression task, we try to predict a precise numerical value based on inputs. Therefore, data augmentation techniques used for regression are not as numerous as those used for classification problems. In the literature, injecting small random noise to the input of a neural network is the most used data augmentation technique for regression \cite{sietsma1991}. Artificial training data is obtained by adding random noise to the original input. Then, these original and artificial data are concatenated. In addition to avoiding overfitting, this technique improves the robustness of neural networks \cite{goodfellow2016}. Based on this idea, we present two data augmentation techniques for predicting nuclear binding energies as well as making a contribution to the techniques used for regression.  

\subsubsection{Data augmentation using the experimental uncertainties}
In our first approach, we apply data augmentation using the experimental uncertainties. We use the experimental data for the relevant observables and their uncertainties. The training data set is resampled twice by adding and subtracting the error values from the original data, while the proton ($Z$) and mass ($A$) numbers are kept constant in the training set. For instance, the experimental data for the total binding energy of $^{208}$Pb is obtained as $1636.43022\pm0.00125$ MeV (see Ref.\cite{Wang_2017}). By using the simple error augmentation technique, the new data for $^{208}$Pb are obtained as $1636.43022$, $1636.43147$, and $1636.42897$ MeV. The same method is used for each nucleus by using their experimental error values. For nuclei with zero error values in their total binding energies, we do not use data augmentation and these nuclei have single data in the training set. Using the experimental data of 1685 nuclei, the training data set is resampled twice and increased to 4995.

\subsubsection{Data augmentation using Gaussian noise distribution} \label{GN}
As an alternative to the data augmentation technique given above, we also use the normal or Gaussian probability density function, which is also known as the bell-shaped probability density function. The Gaussian probability density function is defined as

\begin{equation}
 f(x) = \frac{1}{\sqrt{2\pi\sigma^2}}\,\mathrm{exp} \left(-\frac{(x-\mu)^2}{2\sigma^2}\right),
 \label{eq:randnormal}
\end{equation}
where $\mu$ and $\sigma$ are the mean and the standard deviation, respectively. In the Gaussian noise resampling, we followed the same procedure as simple error augmentation. The proton ($Z$) and mass ($A$) numbers are kept constant and the training data set is resampled by creating random samples from Eq.~\ref{eq:randnormal}. We augmented the training data up to 5 resamplings using this technique. We save the increased data every time we resample and add the next resample over previous data. The fluctuation that comes from randomness is minimized using this resampling method since each sample contains the data of the previous one.

\begin{figure}[h]
\centering
\includegraphics[width=0.8\linewidth]{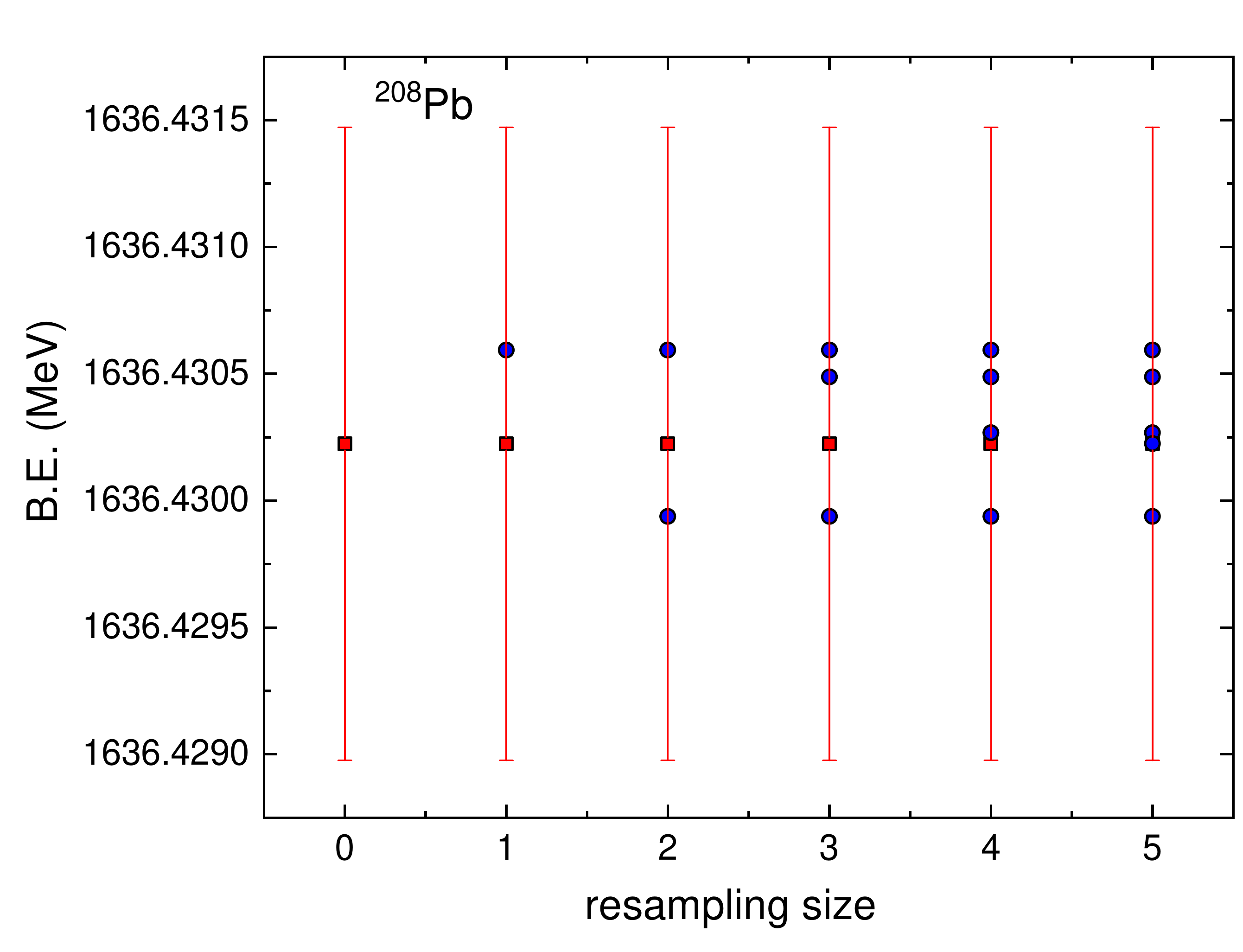}
\caption{Experimental value for the total binding energy of $^{208}$Pb (red square) with the corresponding error value (red line). Application of the Gaussian noise distribution technique to increase the total binding energy data for $^{208}$Pb using different resampling sizes (blue circles).}
\label{fig:208Pb}
\end{figure}

To illustrate the working mechanism of the Gaussian augmentation technique, we use $^{208}$Pb as an example. The experimental data for $^{208}$Pb is given as $1636.43022\pm0.00125$ MeV \cite{Wang_2017}. Here, the mean deviation is given as $\mu=1636.43022$, and the standard deviation is $\sigma = 0.00125$. We resampled the $^{208}$Pb according to these values using the Eq.~\ref{eq:randnormal} and got binding energy for the resampled $^{208}$Pb as $1636.43059$ MeV, we resampled again and got the binding energy for the second resample as $1636.4299$ MeV. We continue to increase the data up to 5 resamples for every value in the training set. Resampling process for $^{208}$Pb is illustrated in Figure ~\ref{fig:208Pb}.

\section{Results and Discussion}

\subsection{Error augmented data}
In this section, we perform our calculations using the MLP architectures with different depths. In the calculations, we use the experimental information of the total binding energies of nuclei with $Z, N\geq$8 (2408 nuclei) from AME2016 \cite{Wang_2017}. Then, the data set is split into training (\%70.0, 1685 nuclei) and test sets (\%30.0, 723 nuclei), as usual. We only use the proton ($Z$) and mass ($A$) numbers of the relevant nuclei as the input and the total binding energies are predicted in the output. After finding the optimum number of epoch and batch for each neural network architecture, 10 calculations are performed using different seed values and the mean values of the r.m.s. errors ($\sigma_{rms}^{Exp.}$) between the test set nuclei and the experimental data are given in Table \ref{table00}. As can be seen from Table \ref{table00}, the different architectures predict similar r.m.s. error values ranging between 1.6 and 2.0 MeV. These findings are also comparable with the results of the modern nuclear energy density functionals \cite{Erler2012}.

Then, the training data set is augmented using the error values in the experimental data of nuclei. To this aim, proton ($Z$) and mass ($A$) numbers are fixed, and artificial data is created by adding and subtracting the error values from the original data set. After data augmentation, training data is resampled twice (excluding nuclei with zero error) and the total number of nuclei is increased to 4995 in the training set. On the other side,  the test set (723 nuclei) is fixed to study the changes in the performance of the neural network models after data augmentation. We perform 10 calculations by choosing the same epoch, batch, and seed values for each architecture. The mean values of the r.m.s. errors ($\sigma_{rms}^{Augmented}$) are given in Table \ref{table00}. After applying data augmentation to the training data set, the performance of some neural networks increases, while some of them are found to decrease.

\begin{table*}
	\caption{The root-mean-square error values ($\sigma_{rms}$, in units of MeV) of total binding energies between the model predictions and experimental data. Different MLP architectures are used in the calculations. The presented values represent the mean values of r.m.s errors obtained using 10 different seed values in the calculations. The last column is the percentage change in the error values after applying data augmentation. We only use proton ($Z$) and mass ($A$) numbers of nuclei as the input. See the text for details.} 
	\begin{center}
		\tabcolsep=0.3em \renewcommand{\arraystretch}{1.0}%
		\begin{tabular}
			[c]{cccccccc}\hline\hline 
			\\ [-1ex]
			MLP& \# of Parameters & Epoch & Batch & $\sigma_{rms}^{Exp.}$  & $\sigma_{rms}^{Augmented}$ & $\%$    \\
			 &    &  &  &1685 nuclei &4995 nuclei  &  \\	\hline
			(128) & 513 & 4500 & 32 & 1.903  & 1.591 & 16.395 \\            
			(32-32) & 1185 & 6000 & 64 &1.773 &1.544 &12.915\\            
			(64-16) & 1249 & 4500 & 32  & 1.678 & 1.647 & 1.847\\           
			(32-32-8) & 1425  & 5500 & 32  &1.760 & 1.592 & 9.545 \\       
			(32-16-8) & 769 & 3500 & 64  & 1.882 &2.022 & $-$7.438   \\           
			(64-16-8) & 1377  & 4500 & 32& 1.865 &1.683 & 9.758  \\      
			(32-16-8-4) & 801 & 7000 & 32& 1.667 & 1.738 &  $-$4.259\\       
			(32-16-16-8) & 1041 &  3500 & 64&  2.147 & 2.081 & 3.074\\    
			(32-32-8-8) & 1497  & 3500 & 32&  2.105 & 1.858 &  11.733  \\     
			(64-16-8-4) & 1409  & 5000 & 64 & 1.843 & 1.976  &$-$7.216  \\    
			\hline\hline
			\end{tabular}\\ [-1ex]
			\end{center}
			\label{table00}
			\end{table*}
			
\subsection{Gaussian noise augmented data}

\begin{figure*}[h!]
\centering
\includegraphics[width=\linewidth]{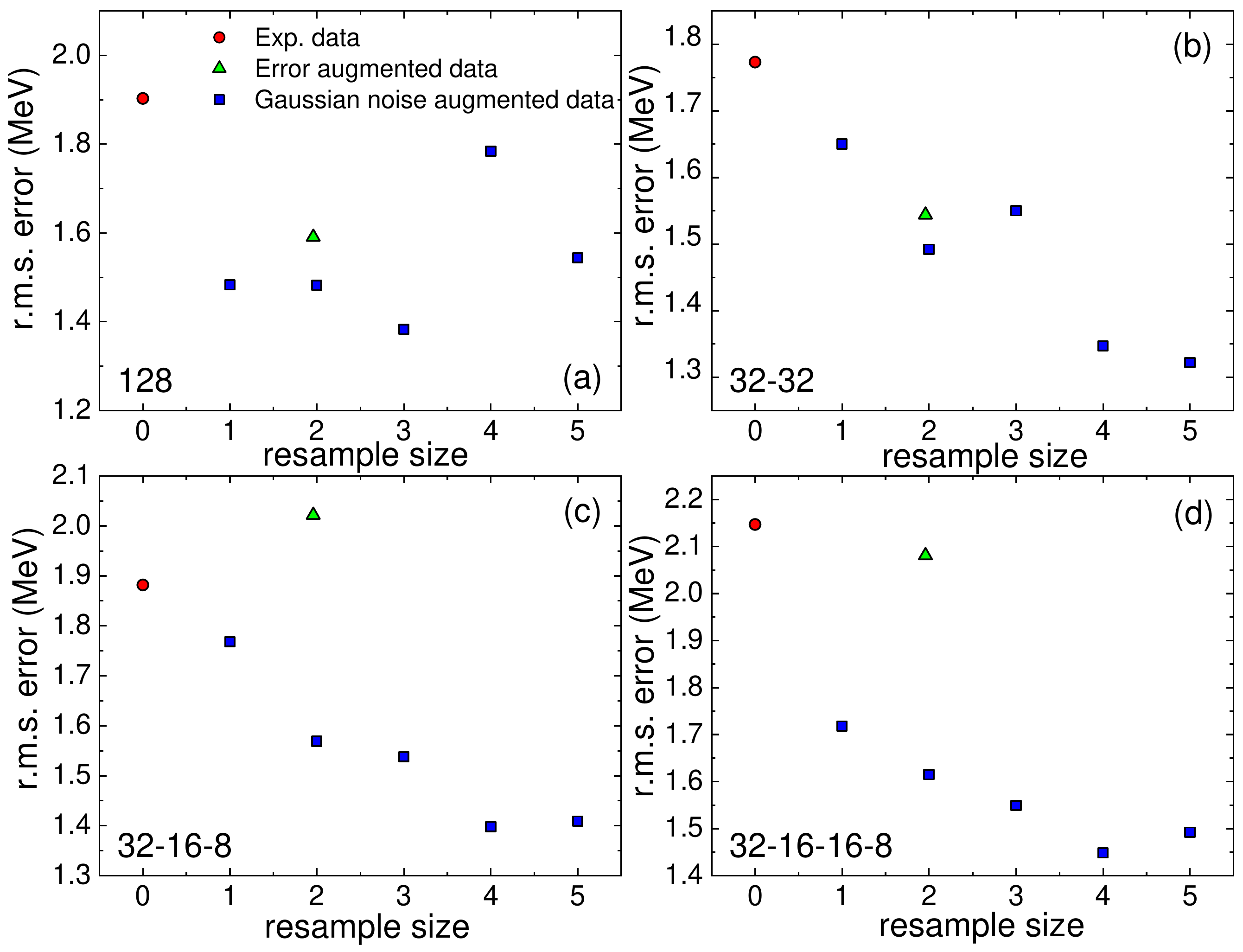}
\caption{The root-mean-square error values as a function of resampling size for (a)128, (b)32-32, (c)32-16-8, and (d)32-16-16-8 MLP architectures.}
\label{rms}
\end{figure*}

In this part, we study the changes in the predictive power of the MLP architectures by increasing the training data set further using the Gaussian noise augmentation technique. As explained in Sec.\ref{GN}, the training data set is augmented by resampling data in different magnitudes. The test set is fixed to study the impact of the data augmentation on the predictions. The calculations are performed using the same epoch, batch and seed values, and the mean values of the r.m.s. errors are obtained. In figure \ref{rms}, we compare the performances of the different neural networks on the test set nuclei by increasing the training data set using the Gaussian noise augmentation technique up to 5 resampling. Using only experimental data in the calculations, the training data set contains 1685 nuclei. After the application of the Gaussian noise technique, the number of the training data set in increased to 3370, 5055, 5740, 8425, and 10110 for 1, 2, 3, 4, and 5 resamplings, respectively.

\begin{figure*}[h!]
\centering
\includegraphics[width=0.8\linewidth]{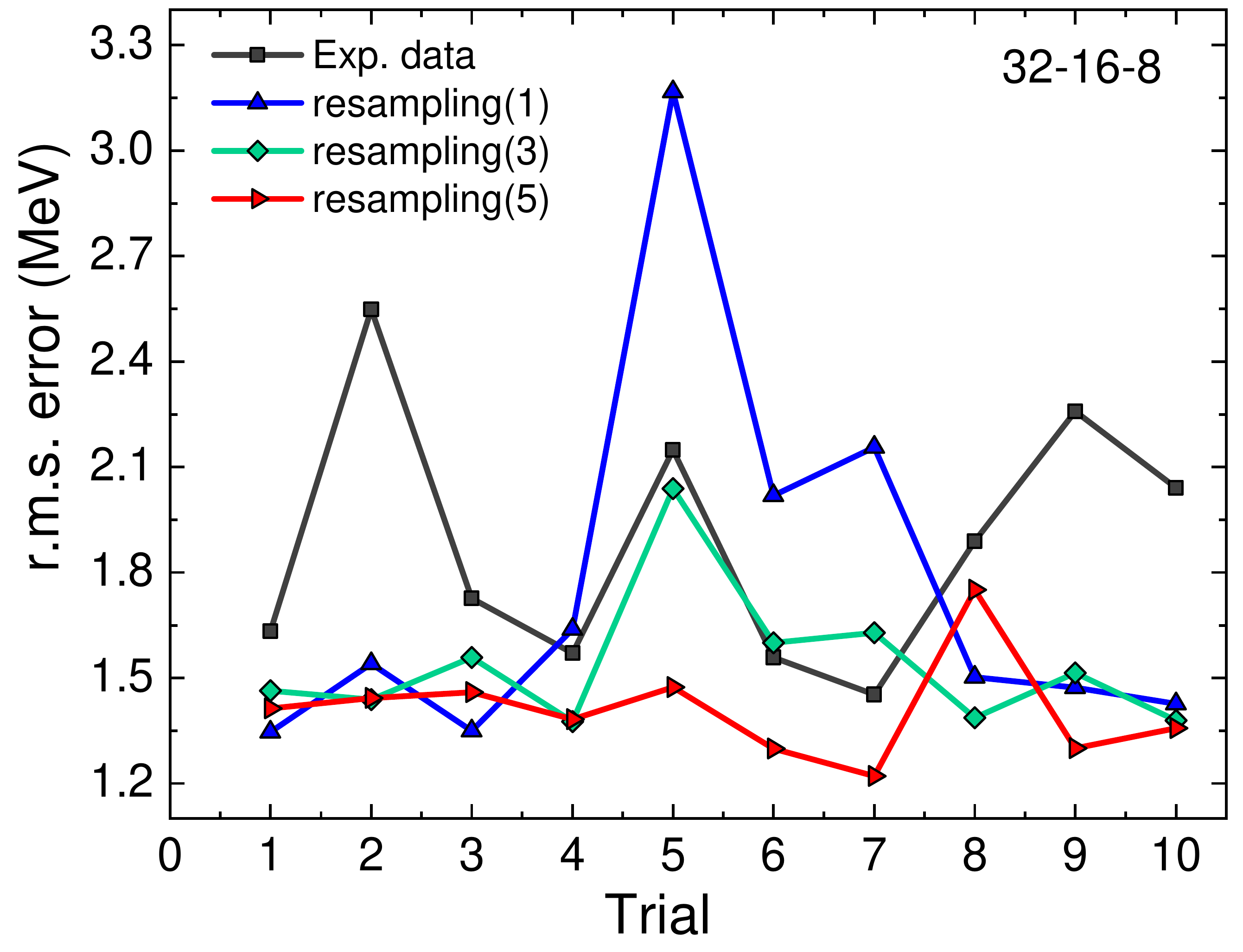}
\caption{The root-mean-square error values using 10 different seed values for 32-16-8 MLP architecture. Three different resampling sizes are used in the calculations.}
\label{stab}
\end{figure*}

It is seen that the performances of neural networks improve by increasing the training data set. In our work, we obtain a gradual increase in the performance of the neural networks up to the 5 resampling for the training data set. For 5 resampling of training data set, the predictive power of the neural networks increases about $\%18.86$, $\%$25.44, $\%$25.12, $\%$30.50 for 128, 32-32, 32-16-8, and 32-16-16-8 architectures, respectively. By increasing the training data set further, we find that the performances of the neural networks do not improve anymore, and we only obtain slight changes. Similar results are obtained for other neural networks (see Table \ref{table00}), and the r.m.s. errors found to decrease with increasing the training data set. Although we obtain fluctuations in the r.m.s. error values for single layer MLP architecture with 128 hidden units, data augmentation gives better results by increasing the number of hidden layers, i.e., for deeper neural networks.

In Figure \ref{stab}, the r.m.s. errors are presented for 32-16-8 architecture for 10 different seed values. To understand the effect of the resampling on the calculations, we display the results for different training data sets, i.e., using only experimental data and Gaussian augmentation in different magnitudes. It is clear that, by increasing the training data with the Gaussian noise augmentation, the fluctuations in the r.m.s. error predictions decrease considerably and the model predictions become more stable against the changes in seed values. According to our findings, data augmentation increases the performance of the neural networks, prevents overfitting, and decreases the variance of the model calculations in different random seed values.

\begin{table*}
	\caption{The root-mean-square error values ($\sigma_{rms}$, in units of MeV) of total binding energies between the model predictions and experimental data. The number of nuclei in the training data set is also provided below the $\sigma_{rms}$. The training data is augmented using the Gaussian Noise augmentation with different resampling sizes. The presented values represent the mean values of r.m.s errors obtained using 10 different seed values in the calculations.} 
	\begin{center}
		\tabcolsep=0.3em \renewcommand{\arraystretch}{1.0}%
		\begin{tabular}
			[c]{cccccccc}\hline\hline 
			\\ [-1ex]
			MLP&  $\sigma_{rms}^{exp.}$  &  $\sigma_{rms}^{1}$  &  $\sigma_{rms}^{2}$  & $\sigma_{rms}^{3}$  & $\sigma_{rms}^{4}$ & $\sigma_{rms}^{5}$     \\
			&1685    & 3370 &5055 &6740  & 8425 & 10110   \\	\hline
			(128) & 1.903 & 1.483 & 1.482 &1.383 & 1.784 & 1.544 \\            
			(32-32) & 1.773 &1.650 &1.492 &1.550 &1.347 & 1.322 \\             
			(64-16) & 1.678 & 1.674 & 1.489 & 1.571 & 1.477 & 1.363\\          
			(32-32-8) &1.760 & 1.424 & 1.607 & 1.434 & 1.614 & 1.426 \\        
			(32-16-8) & 1.882 &1.761 & 1.570 & 1.538 & 1.398 & 1.409  \\       
			(64-16-8) & 1.865 &1.527 & 1.468 & 1.580 & 1.383 & 1.426  \\       
			(32-16-8-4) & 1.667 & 1.639 & 1.511 & 1.583 & 1.526 &1.463 \\      
			(32-16-16-8) & 2.147 & 1.718 & 1.615 & 1.549 & 1.448 & 1.492 \\    
			(32-32-8-8) & 2.105 & 1.616 &  1.709 & 1.806 &1.662 & 1.728  \\    
			(64-16-8-4) & 1.843 & 1.586  & 1.550 & 1.440 & 1.586 &1.324 \\     
			\hline\hline
		\end{tabular}\\ [-1ex]
	\end{center}
	\label{table01}
\end{table*}

\subsection{Extrapolation for the new nuclei using MLP}
While neural networks perform well on the test set within the training data region, the same is not valid for the extrapolation, i.e., learning outside the training data set. The performance of the neural networks decreases considerably outside the training data set. Nevertheless, it would be interesting to study the changes in the extrapolation capabilities of the neural networks after the training data set is augmented artificially. Our question is “Does data augmentation affect the extrapolation performance of neural networks?”

\begin{figure*}[ht!]
	\centering
	\includegraphics[width=\linewidth]{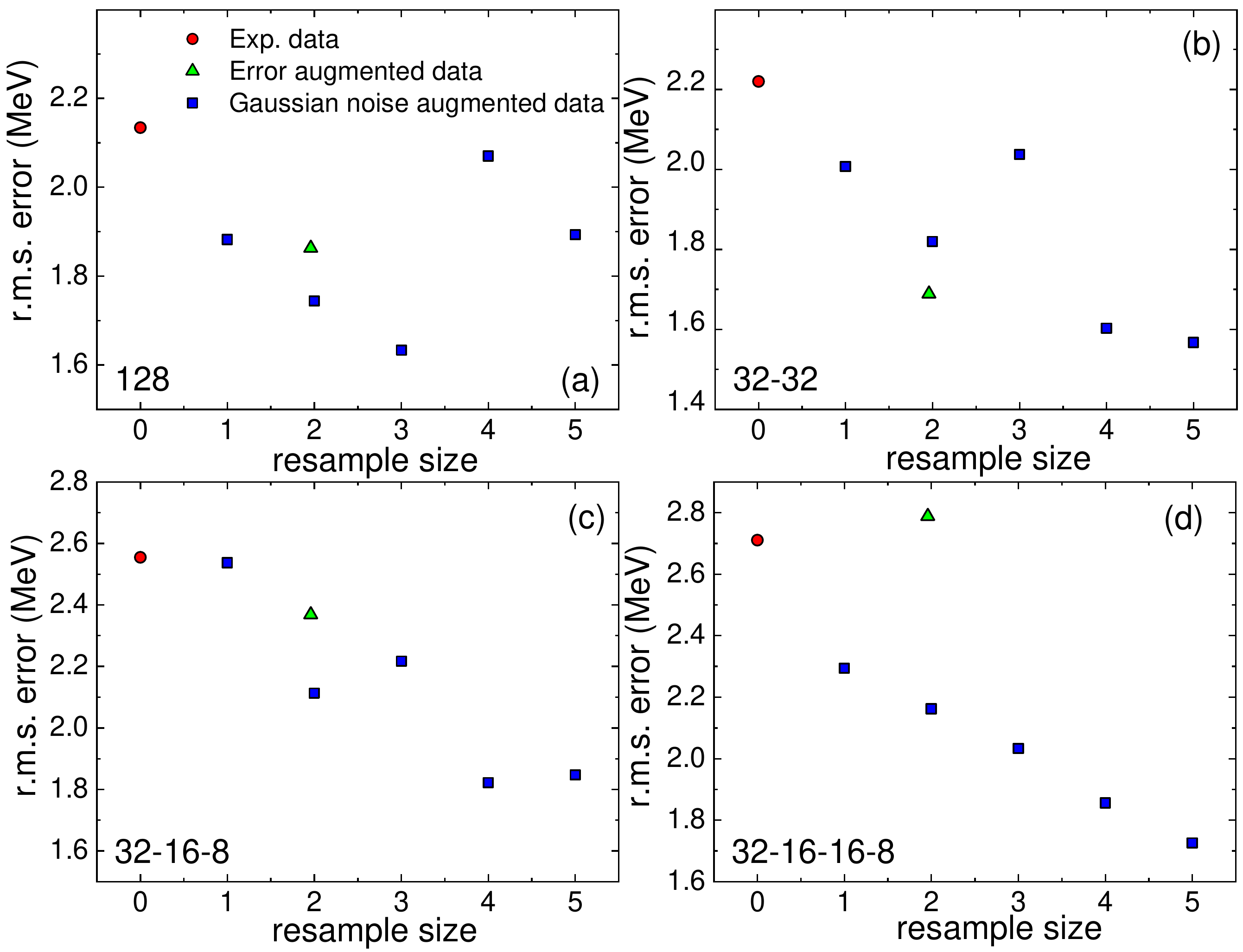}
	\caption{The same as in Figure \ref{rms}, but for the new nuclei in AME2020 table \cite{Wang_2021}.}
	\label{ext}
\end{figure*}

\begin{table*}
	\caption{The same as in Table \ref{table01}, but for the new nuclei in AME2020 table \cite{Wang_2021}.}
	\begin{center}
		\tabcolsep=0.3em \renewcommand{\arraystretch}{1.0}%
		\begin{tabular}
			[c]{cccccccc}\hline\hline 
			\\ [-1ex]
			MLP&  $\sigma_{rms}^{exp.}$  &  $\sigma_{rms}^{1}$  &  $\sigma_{rms}^{2}$  & $\sigma_{rms}^{3}$  & $\sigma_{rms}^{4}$ & $\sigma_{rms}^{5}$     \\
			&1685    & 3370 &5055 &6740  & 8425 & 10110   \\	\hline
			(128) & 2.134 & 1.882 & 1.744 &1.633 & 2.069 & 1.893 \\            
			(32-32) & 2.220 &2.00 &1.820 &2.037 &1.603 & 1.567 \\            
			(64-16) & 2.00 & 2.127 & 1.866 & 1.879 & 1.819 & 1.697\\           
			(32-32-8) &2.055 & 1.677 & 1.934 & 1.626 & 1.938 & 1.825 \\       
			(32-16-8) & 2.555 &2.537 & 2.113 & 2.217 & 1.822 & 1.847 \\           
			(64-16-8) & 2.336 &1.834 & 1.766 & 1.890 & 1.751 & 1.831 \\      
			(32-16-8-4) & 2.089 & 1.943 & 1.870 & 1.987 & 1.846 &1.972 \\       
			(32-16-16-8) & 2.711 & 2.294 & 2.162 & 2.033 & 1.856 & 1.726 \\    
			(32-32-8-8) & 2.433 & 1.861 &  1.918 & 2.092 &1.983 & 2.121  \\     
			(64-16-8-4) & 2.154 & 1.808  & 1.780 & 1.581 & 1.797 &1.495 \\    
			\hline\hline
		\end{tabular}\\ [-1ex]
	\end{center}
	\label{table02}
\end{table*}

Recently, the new Atomic Mass Evaluation - AME2020 table has been published (see Refs. \cite{Huang_2021,Wang_2021}). Compared to the AME2016 set, the binding energies of 71 new nuclei have been announced. These nuclei are mainly placed close to the neutron drip line with extreme proton-to-neutron ratios, therefore they are suitable candidates to check the success of our models. Using these nuclei in the predictions, we test the extrapolation capabilities of neural network models and the effect of data augmentation on the predictions. Using the same epoch and batch values in the MLP architectures (see Table \ref{table00}), 10 different calculations are performed with different seed values and the mean values of the r.m.s. errors are calculated.

In Figure \ref{ext}, we display the mean values of the r.m.s. errors between the MLP calculations and the experimental data as a function of resample size. Without augmentation and using only the experimental data in the training set, the r.m.s. errors are high for the extrapolation region and obtained above 2.2 MeV for each MLP architecture. Similar to the results obtained in Figure \ref{rms}, the r.m.s. error values decrease gradually with increasing training data set.  Although this decrease is not smooth for all MLP architectures, the predictive power of the MLP in the extrapolation region increases considerably. After augmentation, the performances of the 128, 32-32, 32-16-8, 32-16-16-8 architectures increase at most by 
$\%$23.47, $\%$29.40, $\%$28.68, $\%$36.33, respectively. We conclude that the application of the data augmentation technique increases the performance of the MLP also for the new nuclei in AME2020 table \cite{Wang_2021}.

\begin{figure*}[h!]
\centering
\includegraphics[width=0.8\linewidth]{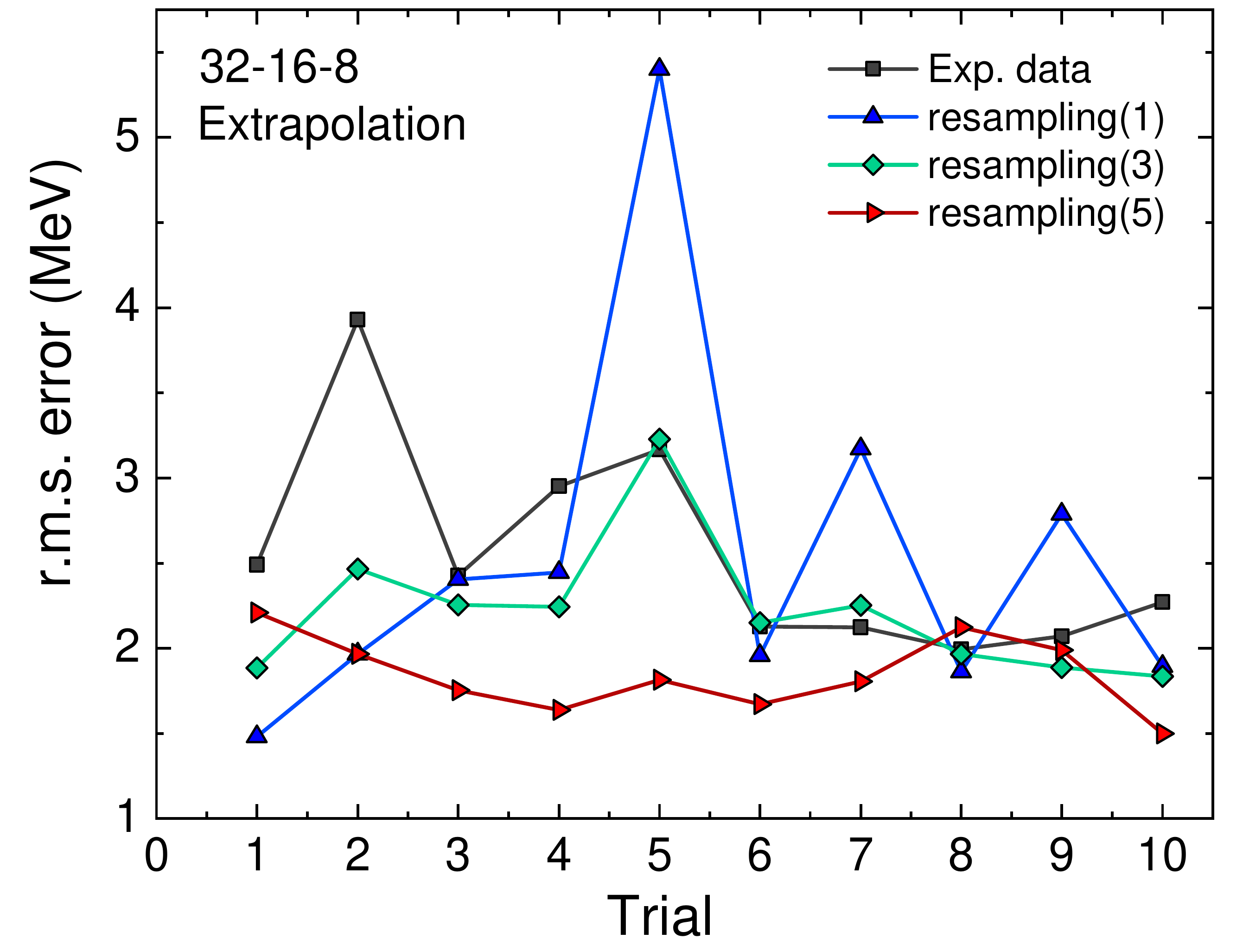}
\caption{The same as in Figure \ref{stab}, but for the new nuclei in AME2020 table \cite{Wang_2021}.}
\label{stab2}
\end{figure*}

In Figure \ref{stab2}, the r.m.s. errors for 10 different calculations with different seed values are displayed for the extrapolation nuclei using the 32-16-8 architecture as a function of the resampling size. Similar to the findings in Figure \ref{stab}, the variance of the model calculations against the changes in the seed number decreases considerably with increasing resampling size in the training data set. Increasing the training data with data augmentation techniques, the stability of neural network architectures also increases for the extrapolation region.

\subsection{Dependence of the results on the optimizer and activation function}
\label{optimizer}

After investigating the effect of the data augmentation on the predictive power of the neural network models, it is also relevant to study the dependence of these findings on the optimizer and activation function. Up to now, the calculations are performed using the Adam algorithm as the optimizer and Rectiﬁed Linear Unit (ReLU) for activation function. In this part, we also use other optimizers commonly used for neural networks to compare our results: Nadam \cite{dozat2016}, AdaMax \cite{kingma2014} and RMSProp \cite{Tieleman2012}. While Nadam and AdaMax are simply modified versions of Adam, RMSProp is an adaptive gradient method proposed before the Adam algorithm. Detailed information about these optimizers can be found in \cite{soydaner2020}. Then, the calculations are also repeated using different activation functions: Tanh and Sigmoid.

\begin{figure}[ht!]
	\centering
	\includegraphics[width=\linewidth]{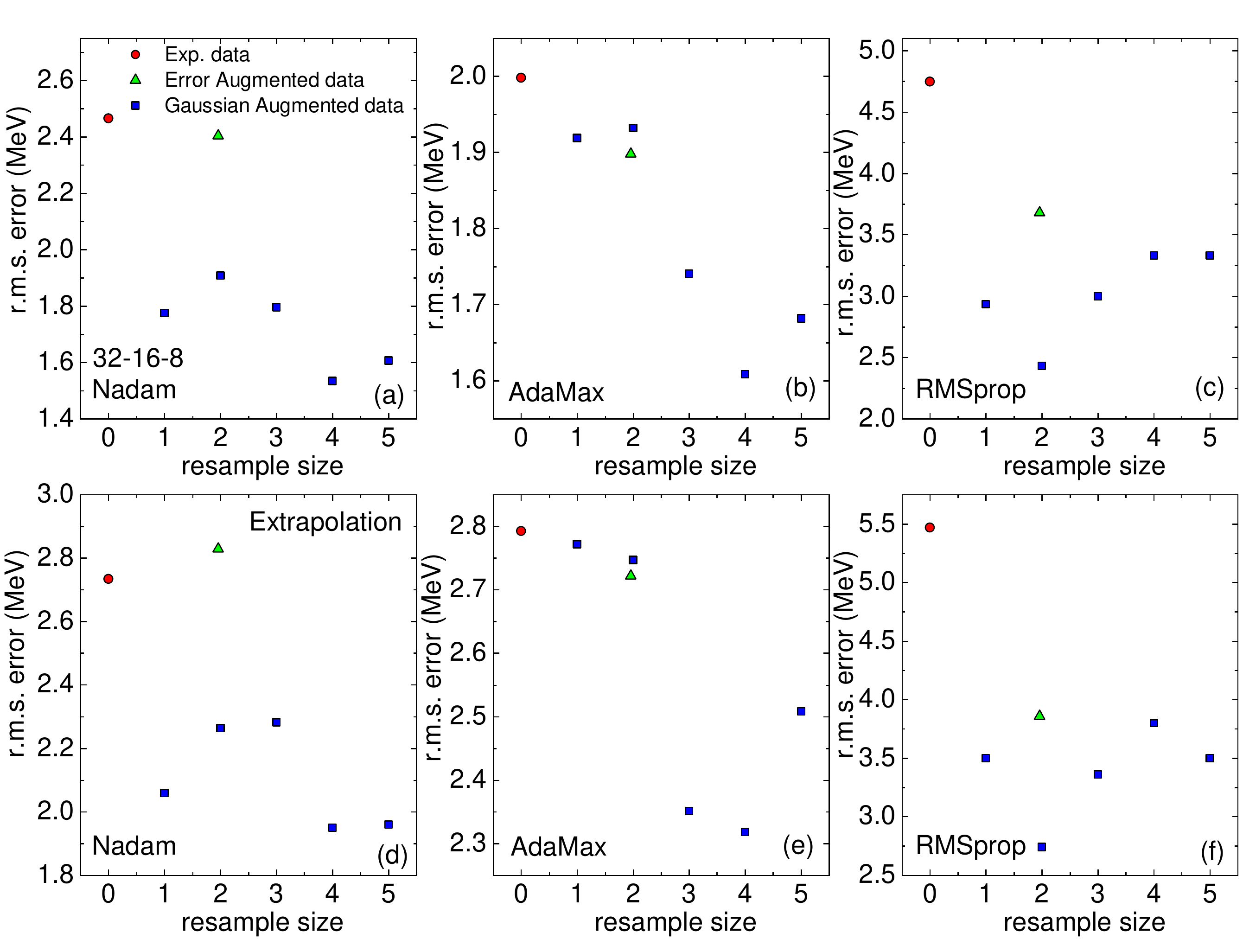}
	\caption{Upper panels: the mean values of the r.m.s. error as a function of resampling size. The calculations are performed using (a) Nadam, (b) AdaMax, and (c) RMSProp as optimizers. Lower panels: The same but for the new nuclei in AME2020 table \cite{Wang_2021} using (d) Nadam, (e) AdaMax, and (f) RMSProp as optimizer.}
	\label{opt}
\end{figure}

In the upper panels of Figure \ref{opt}, we display the r.m.s. error values on the test set nuclei by using three different optimizers in the calculations, and the ReLU is used as the activation function. To test the impact of the optimizer on the results, we select 32-16-8 MLP architecture.  Comparing Nadam, AdaMax, and RMSProp, the best results are obtained using the Nadam and AdaMax, while the RMSProp makes the worst predictions for the test set nuclei. Similar to the findings above, the predictive power of the MLP architecture increases considerably by increasing the training data set with augmentation. For instance, the highest improvements in the r.m.s. error values are obtained as \%37.80, \%19.46, and \%48.80 for the Nadam, AdaMax, and RMSProp optimizers, respectively. It is seen that data augmentation enhances the predictive power of the MLP models with low predictive abilities.
In the lower panels of Figure \ref{opt}, we also display the r.m.s. error values for the extrapolation region, i.e., for new nuclei in AME2020 mass table \cite{Wang_2021}. The same trend is also obtained for this region, i.e., the r.m.s. error decreases with increasing training data set using augmentation technique. After data augmentation, the improvement in the results reaches \%28.70, \%17.0, and \%50.0 for Nadam, AdaMax, and RMSProp optimizers, respectively.

\begin{figure}[ht!]
	\centering
	\includegraphics[width=\linewidth]{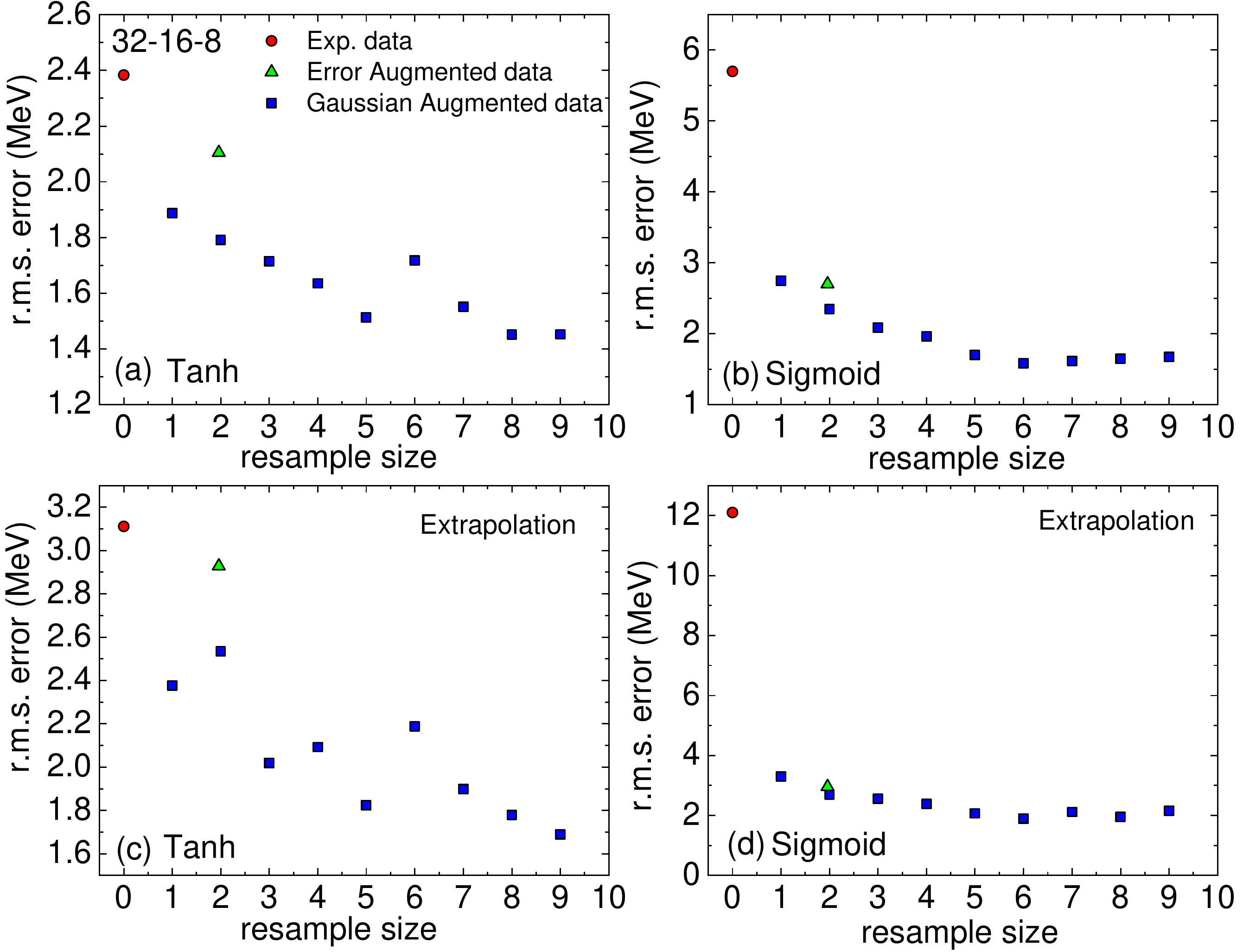}
	\caption{Upper panels: the mean values of the r.m.s. error as a function of resampling size. The calculations are performed using (a) Tanh, and (b) Sigmoid as activation functions. Lower panel: The same but for the new nuclei in AME2020 table \cite{Wang_2021} using (c) Tanh, and (d) Sigmoid.}
	\label{act}
\end{figure}

In the upper panels of Figure \ref{act}, the r.m.s. errors for the test set nuclei are shown as a function of resampling size using the Tanh (a) and Sigmoid (b) as activation functions to train the 32-16-8 MLP architecture. Adam algorithm is used as the optimizer and the resampling size is increased further up to 9 (16850 nuclei in the training set). Similar to the findings above, the predictions on the test set nuclei are greatly improved. This improvement reaches \%72.22 for the Sigmoid activation function, while we obtain \%39.10 improvement in the predictions using Tanh as our activation function.
Similar results are also obtained for extrapolation nuclei (see the lower panel of Figure \ref{act}): the predictions are improved by \%45.71 and \%83.90 using Tanh and Sigmoid, respectively. Our findings show that the impact of the data augmentation is quite high for the models with bad predictions, and the predictions are greatly improved after increasing the training data set artificially.

\section{Conclusions}
In this work, we study the effect of data augmentation on the predictive power of neural network models. To this aim, we use the experimental data of total binding energies and the corresponding uncertainties of 2408 nuclei with $Z, N\geq$8. We use both the experimental error values and the Gaussian augmentation techniques to increase the training data set artificially. After finding the optimum epoch and batch values, we performed 10 calculations with different seed values and the mean value of the r.m.s. errors are calculated.

First, we use the Adam algorithm and  ReLU as the optimizer and activation function, respectively. Without data augmentation, the MLP architectures predict similar r.m.s. error values, ranging between 1.6-2.0 MeV. By increasing the training data with error augmentation, some MLP models perform better on test set nuclei. Increasing the training data further and step by step with the Gaussian augmentation technique, it is seen that the performances of the MLP architectures improve. Up to 5 resampling for training data, we obtain an increase in the predictive power of MLP models and the r.m.s. values do not decrease anymore by increasing the training data further. To test the effect of the data augmentation on the extrapolation region, we also calculate the total binding energies of nuclei discovered in the AME2020 mass table. Although the r.m.s. errors are high for these nuclei, the performances of the MLP architectures also improve by increasing training data with augmentation.  We find that data augmentation increases the performance of the neural networks and decreases the variance of the model calculations in different random seed values as well as prevents overfitting.

We also performed calculations using different optimizers and activation functions to check the effects of the data augmentation on the predictions. Similar to the findings above, the performances of the MLP models increase considerably with data augmentation. The results are found to be significantly improved for MLP models with low predictive abilities, including both test and extrapolation nuclei.

As far as we know, these are the first results in nuclear physics involving data augmentation. Our findings show that data augmentation techniques play an important role in the improvement of the MLP models. Besides, data augmentation techniques can be used safely to improve the performance of neural networks not only in nuclear physics data but also for other scientific fields with a small amount of data. Also, one might explore different approaches for data augmentation, such as Markov Chain Monte Carlo (MCMC) methods in Bayesian neural networks. These issues can be the subject of forthcoming works.

\section*{Acknowledgement}
The numerical calculations reported in this paper were partially performed at T\"{U}B\.{I}TAK ULAKB\.{I}M, High Performance and Grid Computing Center (TRUBA resources).

\bibliography{mybibfile}

\end{document}